\documentclass[letterpaper, 10 pt, conference]{ieeeconf}  % Comment this line out if you need a4paper

\IEEEoverridecommandlockouts                              % This command is only needed if 
\usepackage{cite}
\usepackage{amsmath,amssymb,amsfonts}
\usepackage{algorithmic}
\usepackage{graphicx}
\usepackage{textcomp}
\usepackage{xcolor}
\usepackage{multirow}
\usepackage{url}
\def\BibTeX{{\rm B\kern-.05em{\sc i\kern-.025em b}\kern-.08em
    T\kern-.1667em\lower.7ex\hbox{E}\kern-.125emX}}

\title{\LARGE \bf
Conversations with Andrea: Visitors' Opinions on Android Robots in a Museum
}

\author{Marcel Heisler$^{1}$ and Christian Becker-Asano$^{1}$% <-this % stops a space
\thanks{$^{1}$Hochschule der Medien, Stuttgart, Germany
        {\tt\small \{heisler,becker-asano\}@hdm-stuttgart.de}}%
}

\begin{document}
\bstctlcite{IEEEexample:BSTcontrol}

\maketitle
\thispagestyle{empty}
\pagestyle{empty}

\begin{abstract}
The android robot Andrea was set up at a public museum in Germany for six consecutive days to have conversations with visitors, fully autonomously. 
%In contrast to previous comparable field trials, Andrea acted fully autonomously.
No specific context was given, so visitors could state their opinions regarding possible use-cases in structured interviews, without any bias. Additionally the 44 interviewees were asked for their general opinions of the robot, their reasons (not) to interact with it and necessary improvements for future use. The android's voice and wig were changed between different days of operation to give varying cues regarding its gender. This did not have a significant impact on the positive overall perception of the robot. Most visitors want the robot to provide information about exhibits in the future, while opinions on other roles, like a receptionist, were both wanted and explicitly not wanted by different visitors. Speaking more languages (than only English) and faster response times were the improvements most desired. These findings from the interviews are in line with an analysis of the system logs, which revealed, that after chitchat and personal questions, most of the 4436 collected requests asked for information related to the museum and to converse in a different language.
The valuable insights gained from these real-world interactions are now used to improve the system to become a useful real-world application.
\end{abstract}

\section{Introduction}
An android robot's outer appearance is explicitly designed to resemble a human as closely as possible. This is expected to allow such robots to communicate in very human-like, multi-modal ways. Thus, androids have already been tested in various use-cases, where such human-like communication is important, e.g., to provide company and information to older adults \cite{cavallo_not_2022, thalmann_nadine_2021}, for stage performances \cite{choi_design_2011}
%, for psychological studies regarding emotional interactions \cite{sato_android_2022} 
and for job interviews or as receptionists \cite{kawahara_intelligent_2021, baka_social_2022}.

Android robots can be considered a sub-class of the more general class of social robots, which have been employed in museums. A recent review paper \cite{gasteiger_deploying_2021} on social robots in museums found that their main use-case is to serve as guides. Less often social robots are employed in museums to entertain or teach the visitors. According to \cite{gasteiger_deploying_2021} most of the robots were found to be acceptable for use within museums and they identify three main themes from the literature that are important factors for successful human-robot interaction (HRI) in museums: 
\begin{enumerate}
    \item facial expressions
    \item movement
    \item communication and speech
\end{enumerate}
For all of them, but especially for the first and third, android robots provide novel technological opportunities that need to be explored. 

Another survey \cite{webster_public_2022} identified that perceived appropriateness is largely driven by perceptions of the usefulness and emotional skill of the robots.
%, the latter being displayed by their ability to perform rich facial expressions. 
Again, the survey authors found that people find robots in museums rather appropriate for providing information about the exhibits or as tour guides. 

Therefore, it seems reasonable to investigate, how an android robot capable of facial expression and lip-synchronous, autonomous speech performs in the context of a museum in Germany, cf.~Section~\ref{sec:setup}. This was assessed by conducting structured interviews with visitors of the museum, cf.~Section~\ref{sec:approach}, who actively or passively interacted with our android robot Andrea that wore either a long- or short-hair wig and spoke with a male or a female voice. In addition, the conversation content was analyzed. The interpretation of the analysis results, cf.~Section~\ref{results}, are informing our next steps discussed in Section~\ref{sec:discussion}. %But first, let's present related work.

\section{Related work}
\label{related_work}
So far, android robots have been set up in museums either as a piece of art or for scientific studies. While installations by artists like Jordan Wolfson and Goshka Macuga \cite{boris_pofalla_androide_2016} % as well as Louisa Clement \cite{louisa_clement_representative_2021} 
are very interesting per se, this work focuses on more interactive social robots that are used for scientific studies. % in a museum context.

While \textit{Nadine the Social Robot} was set up for five months at the ArtScience Museum in Singapore in 2017 no formal user study was carried out due to a confidentiality agreement \cite{thalmann_nadine_2021}. However, its installation in an office space in 2019 working alongside human co-workers was rated quite positively \cite{vishwanath_humanoid_2019}, although its response time needed improvements and its hands were judged as too big.

%To the best of our knowledge there were three studies previously carried out employing android robots in museums \cite{rosenthal2014uncanny, mara_science_2015, becker-asano_exploring_2010}. All of them took place at the Austrian technology museum Ars Electronica Center or the cafe CUBUS located inside of the museum. 

%The study in the cafe was the first to be carried out (though published second) . 
In Linz, Austria, the android robot Geminoid HI-1 was placed behind a table in the public café on top of the Ars Electronica museum with information material on Japan next to it \cite{rosenthal2014uncanny}. Three conditions were realized: (1) passively watching a laptop in front of it, (2) autonomously looking up at people, who passed by, and (3) being tele-operated. Structured interviews and video analyses revealed that the more eye contact the robot showed, the easier it would be to detect it as a robot. The android was perceived rather as interesting than uncanny. %, though emotional experiences were not asked in this study.

%The second study  took place during the ARS Electronica festival, about one month after the previous study. 
Geminoid HI-1 was later tele-operated during the ARS Electronica festival to interact with visitors \cite{becker-asano_exploring_2010}. Qualitatively analyzed interviews were used to investigate the Uncanny Valley hypothesis \cite{mori_uncanny_2012}. The authors found more positive than negative descriptions of the robot and 37.5\% of the visitors reported an uncanny feeling, while 29\% reported having enjoyed the conversation with it. 
%Visitors were also asked if they could imagine android robots as a medium for tele-conferencing or what other applications might be suitable for such robots. The majority of the visitors were very sceptical about the use for tele-conferencing and the authors did not report about any other applications mentioned, since they considered the answers as too biased by the influence of reports on robot technology in the media.

In order to investigate the Uncanny Valley Effect, museum visitors interacted with a tele-operated \textit{Telenoid} robot, installed in the Ars Electronica Center's public robot lab in 2015 \cite{mara_science_2015}. This robot's appearance is much less human-like than most other android robots. The results show that framing a situational context in terms of a science-fiction story before letting visitors interact significantly reduced the eeriness ratings of the robot.

%conducted experiments with Alter3 at the NRW-Forum, Düsseldorf” (Masumori et al., 2021, p. 7)

Even without a robotic embodiment the large language model (LLM) "ChatGPT 4" has been used recently as core to craft an audio-based museum guide \cite{Trichopoulos2023}. In contrast to traditional, rule-based dialogue systems as, for example, employed in virtual characters \cite{kopp_conversational_2005}, the use of LLMs tends to result in more convincing and eloquent responses. However, the responses of LLM-driven systems are much less controllable and can even be factually wrong.

%Recently, the android robot ERICA \cite{glas_erica_2016} was tele-operated in the lab to collect a corpus of persuasive dialogues containing multimodal information in the aim to create a dialogue system that encourages users to change their behaviors \cite{kawano_multimodal_2022}. A video-based evaluation of ERICA's persuasiveness \cite{augustine2024assessing} has shown that this also depends on the individual human's personality.

%Vorschlag based on recent works - removed to save some space
%More recent works with android robots concentrated on improving their dialog systems, including gestures, for specific tasks or to achieve specific goals \cite{yamamoto_character_2024, kawakubo_asymmetric_2024, uchida_opinion_2024, namba_how_2024} as well the automated and objective evaluation of such embodied dialogue systems \cite{inoue_analysis_2024}. Others conducted research on the role of touch in android robot systems \cite{shiomi_social_2024} or such robots' capabilities to persuade humans \cite{kawano_multimodal_2022, augustine2024assessing}. Lastly technical improvements esp. regarding tele-operation of android robot heads were achieved \cite{nakajima_development_2024}. These recent works are just slightly related to the field study described here, but still their findings may be useful to further improve android Andrea's capabilities.

None of these previous studies let an android robot autonomously talk to museum visitors, and most of them focus on specific aspects like the uncanny valley. To the best of our knowledge, the android robot Andrea presented here is the first to employ an LLM letting it converse autonomously with uninformed visitors of a public museum.
%here the feasibility of employing such an android in a museum is evaluated.

\section{Setup of the empirical study}
\label{sec:setup}

An android robot was employed in a museum as a fully autonomous conversational agent with ChatGPT 3.5 at its core. The interviews, conducted after the visitors had interacted with the robot, focused on their motivations for doing so (or not doing so), their opinions on use-cases of such robots in the museum, as well as suggested improvements. Additionally, this study examined the topics about which visitors conversed with the robot.

\begin{figure}[ht]
    \centering
    \includegraphics[width=0.9\linewidth]{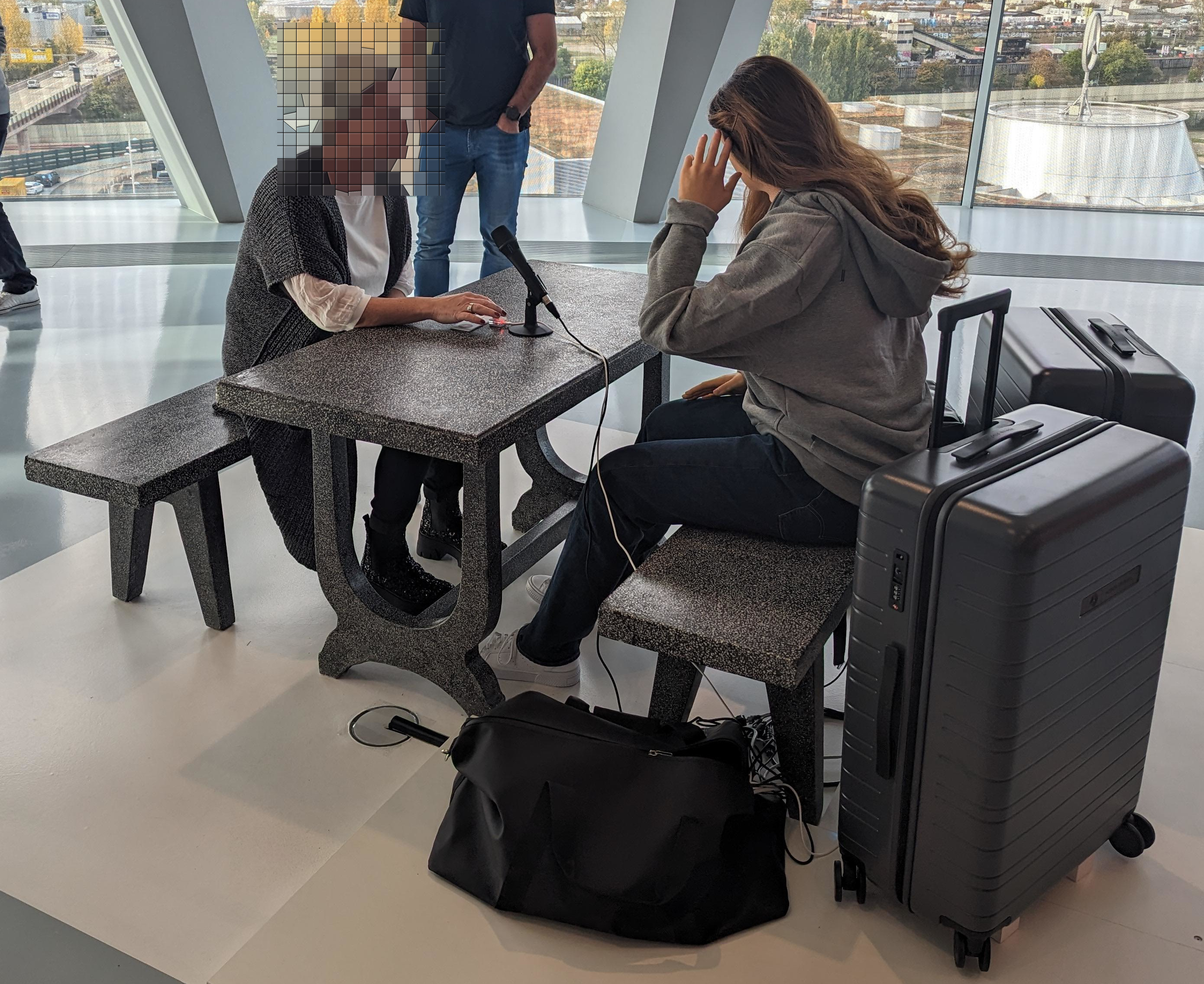}
    \caption{A view from the back of the setup showing the interaction of a visitor through speech with Andrea listening. The weekender travel bag to the left of Andrea contains the computing hardware, whereas the big suitcases behind Andrea are empty and fixed in place to secure the back of the robot.}
    %\Description{A person sitting on the stone bench in front of the android robot. The person is pressing the button on the table between them to interact with the robot.}
    \label{fig:backview}
\end{figure}

\subsection{Hardware}
The hardware of the android robot Andrea was built by the Japanese company A-Lab\footnote{\url{https://www.a-lab-japan.co.jp/en/}}. It is comparable to their \textit{Premium Model} in sitting position with additional actuators to control the movements of its left and right eyelids, as well as each of its finger separately. In total, this results in 52 pneumatic actuators to control the robot's upper body and face.

Its outer appearance is custom designed to feature an androgynous look. A short haired and a long haired wig are available to change the appearance between male and female.
It is dressed casually in blue jeans and a grey zip-up hoodie with a small university logo, cf.~Fig.~\ref{fig:backview} and \ref{fig:animations}.

Cameras are installed in each of the robot's eyes. %, though only the one in the left eye is used to track people, as described in Section~\ref{sec:software}. 
%The robot is extended by external hardware.
A speaker hidden in its hoodie and a microphone with a push-to-talk button on the table in front of the robot enable it to have a conversation. The required software runs on a single Nvidia Jetson Orin placed in a weekender travel bag beside the robot, cf.~Fig.~\ref{fig:backview}, with the robot being connected to it via USB. The robot is also supplied with compressed air (seven bar) and electric power.

\subsection{Software}
\label{sec:software}
The software mainly consists of a pipeline of Machine Learning (ML) models for Speech-to-Text (STT), text-based chat, Text-to-Speech (TTS) and automated lip-sync, as described in \cite{heisler_android_2023}. For STT, \textit{Whisper} \cite{radford_robust_2022} is used by running a \textit{medium} sized model locally on the Jetson Orin. It is used in \textit{translate} mode, so that the model transcribes everything into English independently of the visitor's spoken language. As chat model ChatGPT version 3.5-turbo is used. Since this service was stateless at that time, our software keeps track of the chat history and sends this message stack with every request.
%See Appendix~\ref{app:prompt} for the prompts that were used. 
For TTS, coqui's\footnote{\url{https://github.com/coqui-ai/TTS}} pre-trained model of Variational Inference with adversarial learning for end-to-end Text-to-Speech (VITS) \cite{kim_conditional_2021} is run on the Jetson Orin. Two different voices were manually pre-selected and varied between different days during the study (see also Section~\ref{sec:museum}): \textit{p225} was used as a female and \textit{p234} as a male voice. A lip-sync model \cite{heisler_making_2023} is used to automatically animate the robot's face according to the audio generated by the TTS model. The pipeline is triggered by pressing the push-to-talk-button, which starts recording audio using the microphone. Releasing the button again, stops the recording and starts the transcription of the audio using the STT model.

\begin{figure}[ht]
    \centering
    \includegraphics[width=1\linewidth]{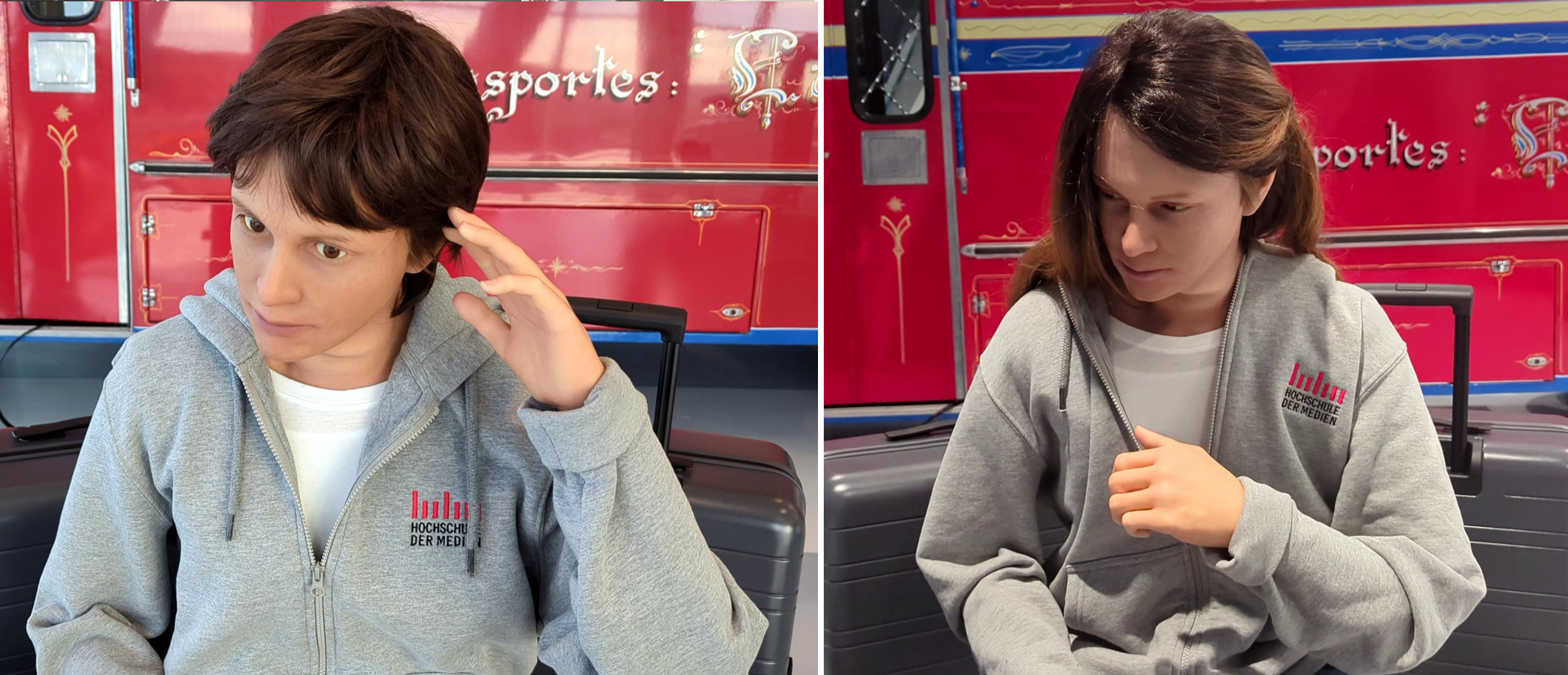}
    \caption{Still images of the two animations: (left) while recording and transcribing the visitor's audio with short-hair wig; (right) while thinking / generating a response with long-hair wig.}
   % \Description{Two (anonymized) images of the android robot Andrea. In the left image its head is turned right so its left ear is turned more forwards and the left hand is close to the ear signaling the robot is listening. In the right image the robot looks downwards squinting its eyes, its left hand is close to its chin, signaling the robot is processing a request.}
    \label{fig:animations}
\end{figure}

%This basic pipeline is accompanied by some manually defined animations as presented by the still images in . 
During audio recording and transcription a \textit{listen} animation is performed by the robot moving its hand to one of its ears, cf.~Figure~\ref{fig:animations} left. During the other steps of the processing pipeline a \textit{thinking} animation is performed, cf.~Figure~\ref{fig:animations} right. Additional animations are executed in random time intervals to resemble blinking, saccades and breathing. If no interaction takes place for a longer interval, the robot yawns.

The video stream of the robot's left eye is analyzed using a pre-trained pose estimation model for Nvidia Jetsons\footnote{\url{https://github.com/dusty-nv/jetson-inference/blob/master/docs/posenet.md}}. The \textit{resnet18-body} model is used to detect 18 keypoints of human bodies in the video stream, from which one nose keypoint is chosen as the target of the robot's gaze. The robot's neck actuators are controlled to center this point in the camera's view. This serves as a rudimentary \textit{look at} animation, which is always active when neither \textit{listen} nor \textit{think} are active.

Currently, there is no automated mechanism in place that distinguishes between different interlocutors. Thus, the current conversation must be reset manually. % by the human operator in the nearby van, cf.~Fig.~\ref{fig:setup-museum}. 
This reset could either be triggered from inside the camper van, cf.~Section~\ref{sec:museum}, or using an extra push-button placed on the floor behind the robot.

\subsection{Museum installation}
\label{sec:museum}

\begin{figure}[ht]
    \centering
    \frame{\includegraphics[width=1.0\linewidth]{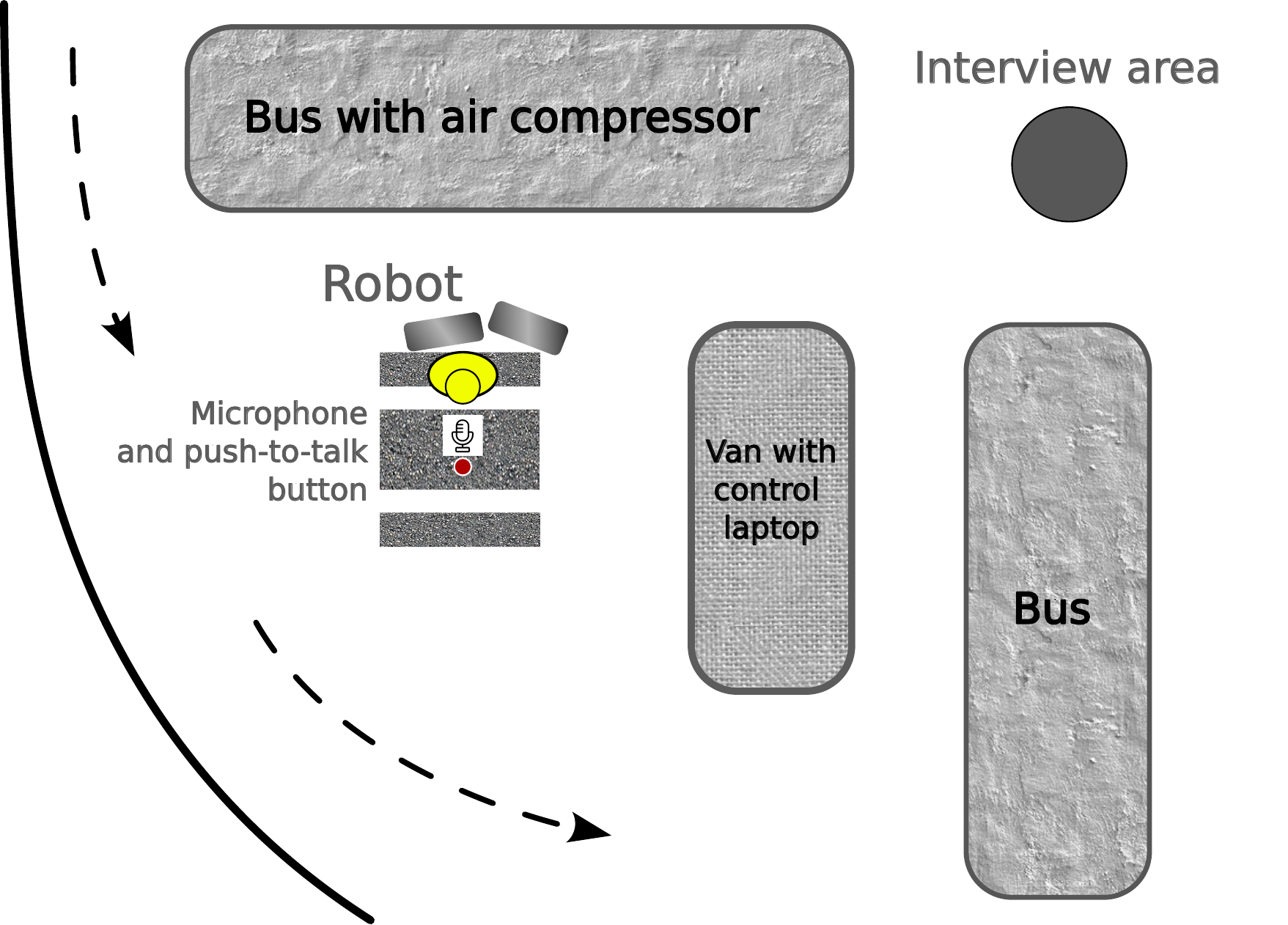}}
    \caption{A sketch of the setup of Andrea during its public demonstration in the Mercedes-Benz museum: The robot is placed on a bench behind a stone table with a microphone and a push-to-talk button on it; the air compressor is hidden in a bus behind the robot; interviews were conducted away from the robot (the dashed arrows indicate the main flow of the visitors)}
    %\Description{The sketch depicts the setup of robot installation in the museum. The robot is placed on a bench with a table in front of it and another bench on the other side of the table. In the back of the robot is a bus where the air compressor is placed in. To left side of the robot the van with the control laptop is parked. Behind the van another bus is depicted. The interview area is a bit offside between the two buses. Visitors came from behind the compressor bus, first seeing the robot from behind.}
    \label{fig:setup-museum}
\end{figure}

The robot was installed for six days---from October 31 to November 5, 2023---in the travel gallery area of the Mercedes-Benz-Museum, cf.~Fig.~\ref{fig:setup-museum}. There was deliberately no advance notice to observe visitors' spontaneous reactions. The robot was ready for interaction during the museum's opening times from 9 am to 6 pm each day.

On the first and fourth day the robot wore a long-haired wig and spoke with its female voice (condition \textit{long-female}). On the second, third, and sixth day it was set up with a short-haired wig and a male voice (condition \textit{short-male}), while on the fifth day it featured a female voice with a short-haired wig (condition \textit{short-female}), cf.~Fig.~\ref{fig:animations}.

Andrea was placed on one of two stone benches that are permanently installed around a stone table for the visitors to take a rest. This way, visitors entering the area from the back did not easily recognize the setup as an artificial installation as indicated by the dashed arrows in Fig.~\ref{fig:setup-museum}. A microphone (Marantz Professional M4U with table stand) and a single push-button were placed in front of Andrea on the table. Visitors were instructed by a small sign placed in front of the button to press and hold it to start the interaction, cf.~Fig.~\ref{fig:backview}. The air compressor was placed in the bus behind the robot and the tube as well as the power supply and a cable connection to the internet were hidden under a fake, wooden floor around the installation. A Mercedes V-class camping van, which is part of the permanent exhibition, was used to supervise the setup.

The interviews with randomly selected visitors were conducted in an area hidden from the interactive installation, cf.~Fig.~\ref{fig:setup-museum}. During their interview the visitors stood with the back to the robot, but could turn around to refer to the setup whenever they liked.

\section{Approach}
\label{sec:approach}

\subsection{Interviews with visitors}
\label{sec:approach-interview}

This setup gave us the opportunity to qualitatively determine Andrea's impact on the visitors through on-location interviews. The open-ended questions were structured as follows:
\begin{enumerate}
    \item Did you interact with the robot?\\
    Yes: continue with (a) and (b); No: continue with (c) and (d)
    \begin{enumerate}
        \item What motivated you to actively interact with the robot?
        \item Please describe your interaction with the robot.
        \item Why did you not interact actively?
        \item Please describe your observations.
    \end{enumerate}
    \item How useful do you think would it be to use this robot in a museum today? (0 - 10)
    \item In which situations would it NOT make sense to use it in the museum? Why not?
    \item In which situations would it be useful to use it in the museum? Why?
    \item What should be urgently improved before such a robot is used?
    \item Do you have any questions and/or suggestions?
\end{enumerate}

After written, informed consent was given, the interviews were conducted in either German or English depending on the visitor's preference. The interviewer was given a handout with the interview structure in English and German, which guided each interview. The voices of both the interviewer and the interviewee were recorded by an audio recorder (TASCAM DR-22WL) and later transcribed for analysis using an offline version of Whisper STT \cite{radford_robust_2022}.

At the beginning of each interview the visitors were asked: (1) for the gender they assigned themselves to, with "male", "female", and "diverse" as options, (2) their age group, with "$< 10$", "$10-20$", "$21-30$", "$31-40$", "$41-50$", "$51-60$", and "$> 60$" as options, and (3) their nationality.
% \begin{itemize}
%     \item for the gender they assigned themselves to, with "male", "female", and "diverse" as options
%     \item their age group, with "< 10", "10-20", "21-30", "31-40", "41-50", "51-60", and "> 60" as options, and
%     \item their nationality.
% \end{itemize}

\subsection{Conversations with Andrea}
\label{sec:approach-conversation}
The software described in Section~\ref{sec:software} logged all interactions during the study on a textual basis, meaning no video or audio recordings were saved, but the transcribed requests from users with the response from ChatGPT were saved for each conversation. Along with input and output texts, the time taken by each of the ML models to process a request, as well as possible errors, were logged. Additionally, resets of the current chat history and whenever a press of the push-to-talk button interrupted the robot's speech were recorded as events. However, interruptions of the \textit{thinking} process were not logged and any information about interruptions was only logged from the third day on.

The log files are analyzed in an automated way. Since the chat history was not always reset after each interlocutor, there is no definitive way to differentiate between different conversations. Thus, all input requests and output responses are extracted separately from the logs by pattern matching.
BERTopic \cite{grootendorst_bertopic_2022} %, a library for topic modelling,
is used to cluster input and output phrases separately.

%BERTopic first creates numerical dense 384-dimensional embeddings for each phrase using SentenceTransformers' \cite{reimers_sentence-bert_2019} \textit{all-MiniLM-L6-v2} model. Next embeddings' dimensionality is reduced to three using UMAP \cite{mcinnes2018umap-software}, then those embedding vectors are clustered using HDBSCAN \cite{McInnes_hdbscan_2017}. These clusters are used to create a bag-of-words matrix, not on the level of individual phrases, but on a cluster level. 
%Class-based term frequency–inverse document frequency is used to extract the most important words for each cluster from this bag-of-words representation. Those words can serve as the cluster names and are also referred to as the extracted topics. In a final step the KeyBert \cite{grootendorst2020keybert} inspired method is applied to fine-tune these topic representations.

After an initial fit to the data, BERTopic's functions to reduce the number of topics and outliers are applied. After subjectively exploring some values, 15 was chosen as a fixed number of topics to reduce to, thus BERTopic applies agglomerative clustering to achieve this reduction. Outliers are reduces using BERTopic's defaults. For each of the remaining 15 clusters (including one for outliers) ten representative phrases are extracted using Maximal Marginal Relevance with a diversity of $0.1$ to avoid very similar representative phrases. % (provided in Appendix~\ref{app:clusters}). 
Finally, those phrases are manually examined to further group the clusters into categories as shown in Section~\ref{sec:conversationAnalysis}.

Apart from user requests and system responses, any errors logged by the system are categorized and counted. Additionally the processing times of the different ML models are analyzed, too.

%\subsection{Social Media}
%might be interesting to compare results from interviews with comments on social media (instagram stz and facebook swr1)?

\section{Results}
\label{results}

\subsection{Exploratory analysis of the interviews}
\label{sec:interviewAnalysis}

\subsubsection{Demographic data}

\begin{table}[ht]
    \centering
    \caption{Distribution of all 44 interviewed visitors depending on gender and whether or not they had actively interacted}
    \includegraphics[width=0.9\linewidth]{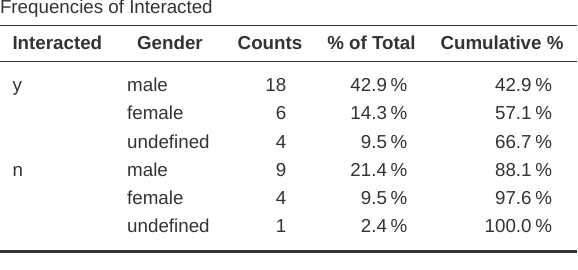}
    \label{fig:table_genderinteracted}
\end{table}

A total of 44 visitors of the museum agreed to be interviewed. They either had directly interacted with Andrea before or had watched Andrea from a distance (i.e., not interacted). For four of them gender, age group, and nationality are missing. One has only gender missing and for two others interaction information is missing.
%out of carelessness of the interviewer.
Table~\ref{fig:table_genderinteracted} gives an overview of the distribution of interviewed visitors. Of the 27 male visitors eighteen had actively interacted with Andrea (67\%) and of the ten female visitors six had interacted (60\%). Of those visitors whose gender is unknown four of the total of five (80\%) had interacted before the interview.

\begin{figure}[ht]
    \centering
    \includegraphics[width=1\linewidth]{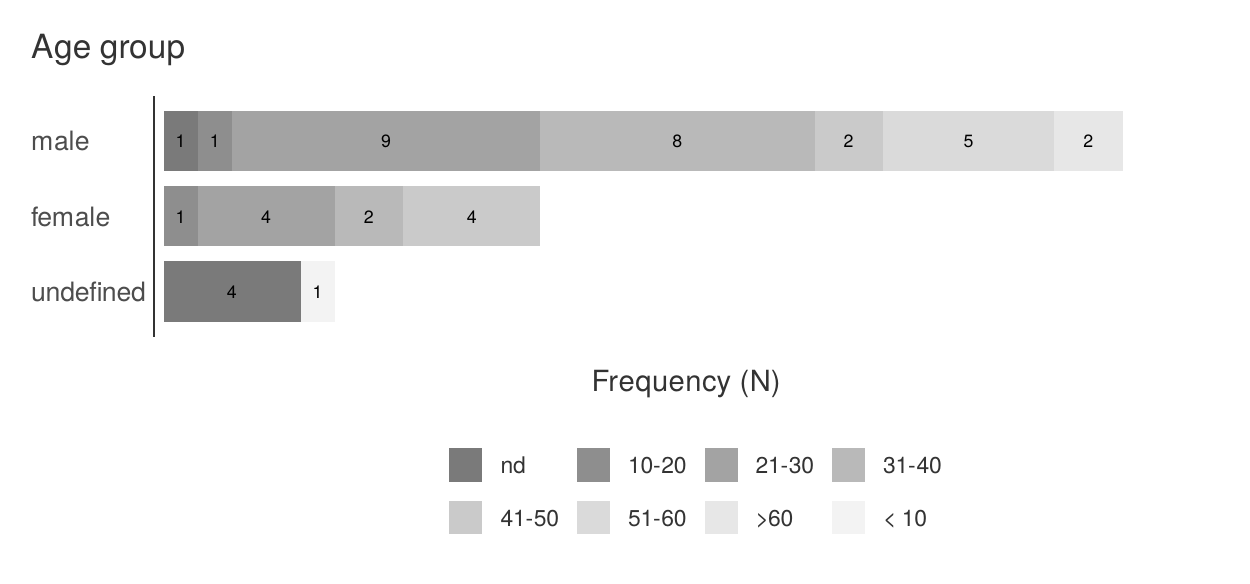}
    \includegraphics[width=1\linewidth]{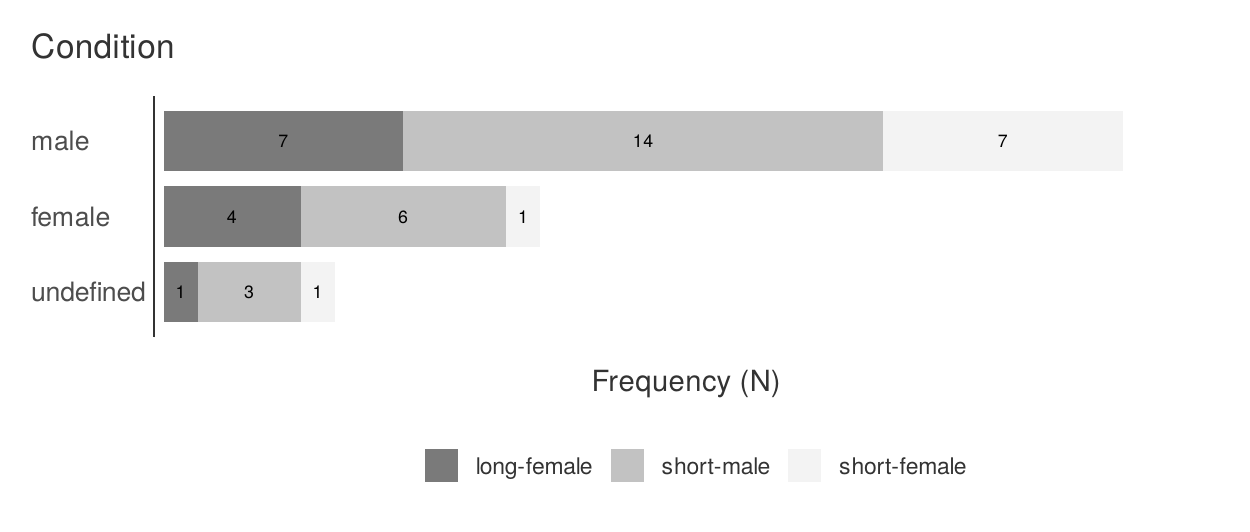}
    \caption{Top: distribution of age groups over gender; bottom: distribution of condition over gender}
    %\Description{Top: Bar chart for age groups and gender. male: nd=1, 10-20=1, 21-30=9, 31-40=8, 41-50=2, 51-60=5, >60=2; female: 10-20=1, 21-30=4, 31-40=2, 41-50=4; undefined gender: nd=4, <10=1.
    %Bottom: Bar chart for robot condition and gender: male: long-female=7, short-male=14, short-female=7; female: long-female=4, short-male=6, short-female=1; undefined gender: long-female=1, short-male=3, short-female=1}
    \label{fig:barcharts_genderdistributions}
\end{figure}

The age group distribution conditioned by gender, cf.~Fig.~\ref{fig:barcharts_genderdistributions}, top, and the experimental conditions split up by gender, cf.~Fig.~\ref{fig:barcharts_genderdistributions}, bottom, show that the interviewees were evenly distributed regarding their respective gender and age over the experimental conditions.

% \begin{figure}[ht]
%     \centering
%     \includegraphics[width=0.6\linewidth]{images/2024-04-12_Table_Nationalities.pdf}
%     \caption{The nationalities of the interviewed visitors.}
%     \label{fig:table-nationalities}
% \end{figure}
Twenty-five of all interviewees said that they were German, another two reported a mixed nationality with German nationality being part of it, and for four interviewees the nationality is unknown. The remaining 15 interviewees are distributed over single countries and, therefore, this factor is not taken into account in the subsequent analysis.

\subsubsection{Interview transcript analysis} The interview structure introduced in Section~\ref{sec:approach-interview} is being followed from here on. First, the automated audio transcripts were manually annotated to distinguish between interviewer and interviewee in each text file. Eventual transcription errors were fixed manually during this step. Then, each transcript was read by the first person to extract main topics per response to each question. For each transcript similarities were checked with previously extracted topics. If topics seemed similar enough, the old topic's count was increased by one. Otherwise, a new topic was introduced with a count of one. The result was critically discussed with a second team member. With this method one interviewee can have contributed to multiple topics within one response. Only those topics with a count of two or more are reported here.

% \begin{itemize}
%     \item[\textbf{(1a+c)}] \textbf{What motivated you to actively interact with the robot? / Why did you not interact actively?}
% \end{itemize}

\paragraph*{(1a+c) What motivated you to actively interact with the robot? / Why did you not interact actively?}

Table~\ref{tab:reason-interact} summarizes the topics, that were extracted from the answers given in response to the question why visitors interacted or not with Andrea before. Two-third of the interviewees (28 out of 42) interacted with Andrea and more than half of these mentioned "curiosity / intrigue" as their motivation. Of those 14, who did not interact, five reported an "uncanny feeling" as a reason.

\begin{table}[ht]
    \centering
    \caption{Reasons to interact or not to interact}
    \begin{tabular}{|l|c|}
        \hline
        Reason to interact & Count\\
        \hline
        curiosity / intrigue & 16\\
        Andrea looking real & 9\\
        general interest in technology / robotics & 5\\
        to test the robot's answers & 5\\
        never before interacted with one & 5\\
        friends motivated me to interact with Andrea & 2\\
        own children interacted & 2\\
        noticed microphone or button & 2\\

        \hline
        Reason not to interact & Count\\
        \hline
        too many people to actively interact & 7\\
        felt uncanny & 5\\
        did not know what to ask (in English) & 3\\
        more information needed / confused & 2\\
        \hline
    \end{tabular}
    \label{tab:reason-interact}
\end{table}

% \begin{itemize}
%     \item[\textbf{(1b+d)}] \textbf{Please describe your interaction with the robot. / Please describe your observations.}
% \end{itemize}

\paragraph*{(1b+d) Please describe your interaction with the robot. / Please describe your observations.}

The descriptions of the interviewed visitors are summarized in Table~\ref{tab:descriptions}. Half of the interviewees mentioned that they wanted to ask some questions and nine of them mistook Andrea for being real at first. Non-verbal interactions by Andrea were mentioned twelve times and, in general, the conversation was mentioned to be too slow and not fluent/natural enough. Four times visitors reported the situation to feel strange, which could be an indication for the uncanny valley effect.

\begin{table}[ht]
    \centering
    \caption{Descriptions of interactions / observations}
    \begin{tabular}{|l|c|}
        \hline
        Description / observation & Count\\
        \hline
        asked (some) questions & 22\\
        Andrea looked real at first, but then... & 9\\
        mentions Andrea's gestures & 7\\
        Andrea tried to make eye contact & 5\\
        conversation is not fluent / natural & 5\\
        strange feeling mentioned & 2\\
        humanoid design is problematic & 2\\
        response time is too long & 2\\
        \hline
    \end{tabular}
    \label{tab:descriptions}
\end{table}

% \begin{itemize}
%     \item[\textbf{(2)}] \textbf{How useful do you think would it be to use this robot in a museum today? (0 - 10)}
% \end{itemize}
\paragraph*{(2) How useful do you think would it be to use this robot in a museum today? (0 - 10)} 

\begin{figure}
    \centering
        \includegraphics[width=0.8\linewidth]{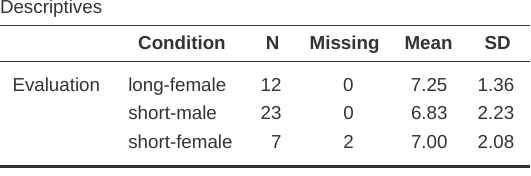}
        \includegraphics[width=0.9\linewidth]{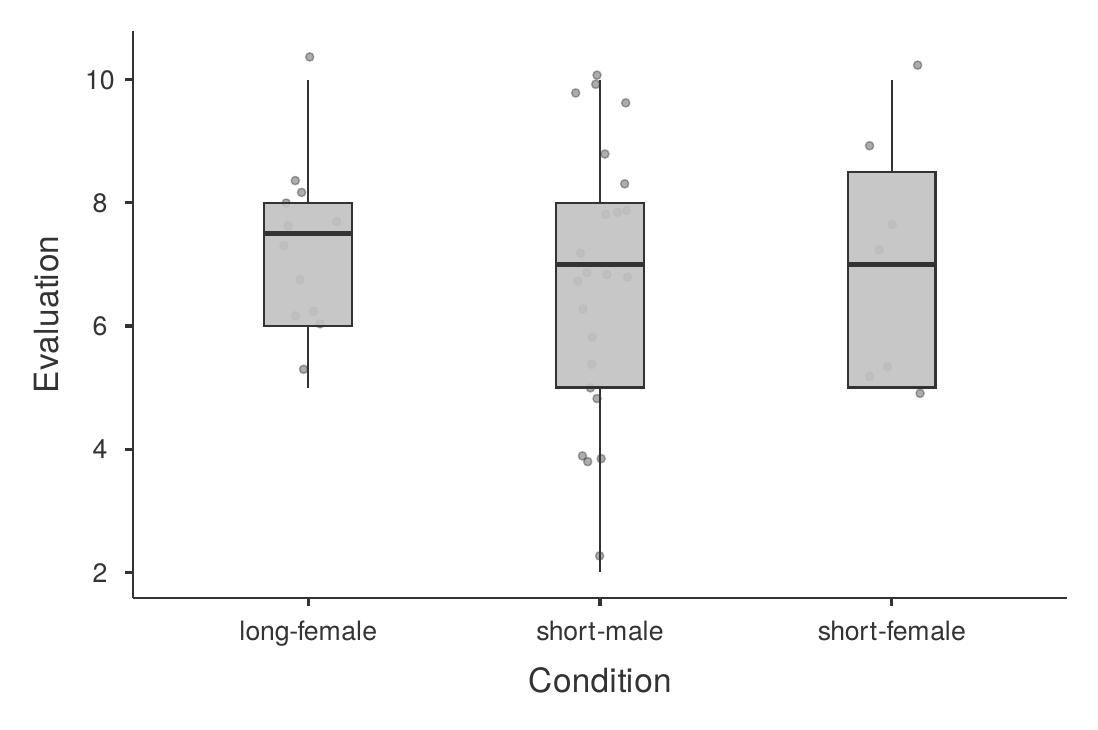}
    \caption{Descriptives of the visitors' evaluations (top) and box-plots (bottom) depending on the type of setup.}
    %\Description{A box plot for each robot condition. long-female: 25\%-75\% between 6 and 8 median between 7 and 8, short-male: 25\%-75\% between 5 and 8 median around 7, short-female: 25\%-75\% between 5 and 9 median around 7}
    \label{fig:boxplot-evaluation}
\end{figure}

The visitors were asked to evaluate the usefulness of Andrea in this situation on a scale from zero to ten. To explore, if any of the three versions of Andrea (long hair and female voice, short hair and male voice, short hair and female voice) are evaluated differently, the ratings are split up accordingly. As the table in Figure~\ref{fig:boxplot-evaluation} (top) shows, twelve interviewees had seen or interacted with Andrea with long hair and a female voice, 23 with short hair and a male voice, and seven with short hair and a female voice. A one-way ANOVA assuming unequal variances with "Condition" as grouping variable results in no significant differences ($F=0.237, df1=2, df2=15.9, p>0.79$). Notably, all mean values of the three conditions are around seven and significantly above the theoretical average of five (pooled evaluation data: Shapiro-Wilk normality test $p=0.056 \rightarrow$ Wilcoxon test with $H_0$ $\mu \neq 5, p<0.001$).

% \begin{itemize}
%     \item[\textbf{(3)}] \textbf{In which situations would it NOT make sense to use it in the museum? Why not?}
% \end{itemize}
\paragraph*{(3) In which situations would it NOT make sense to use it in the museum? Why not?}

Table~\ref{tab:neg-scenarios} presents the topics that were mentioned when the visitors were asked about inappropriate situations for the robot in the museum. Many visitors were surprised to be asked this question and even after thinking about it they replied that no such scenario came to their mind. The best agreement for a specific scenario that Andrea should not be used in was as a simple substitute for the already present audio guides. Some visitors also didn't want the robot to substitute any human in his or her job.

\begin{table}[ht]
    \centering
    \caption{Negative scenarios mentioned by visitors}
    \begin{tabular}{|l|c|}
        \hline
        Negative scenario & Count\\
        \hline
        (no negative scenarios) & 14\\
        presenting information similar to an audio guide & 4\\
        when it substitutes a human & 3\\
        as a guide / during a guided tour & 3\\
        for general question answering & 2\\
        humans should have a choice & 2\\
        whenever asking concrete questions & 2\\
        at the ticket counter & 2\\
        inside/together with old cars & 2\\
        \hline
    \end{tabular}
    \label{tab:neg-scenarios}
\end{table}

% \begin{itemize}
%     \item[\textbf{(4)}] \textbf{In which situations would it be useful to use it in the museum? Why?}
% \end{itemize}
\paragraph*{(4) In which situations would it be useful to use it in the museum? Why?}

\begin{table}[ht]
    \centering
    \caption{Positive scenarios mentioned by visitors}
    \begin{tabular}{|l|c|}
        \hline
        Positive scenario & Count\\
        \hline
        presenting knowledge about exhibits & 14\\
        at the kiosk / ticket counter & 9\\
        instead of audio guides & 7\\
        when people have (more specific) questions & 7\\
        instead of / as a human guide & 5\\
        presenting the theme of the current area & 4\\
        (no positive scenarios) & 4\\
        providing local information (e.g., toilet) & 3\\
        when more info is needed than on display & 3\\
        to present the highlights only & 2\\
        in one-to-one up to one-to-three interaction only & 2\\
        inside the cars & 2\\
        as a supplement to a human guide & 2\\
        \hline
    \end{tabular}
    \label{tab:pos-scenarios}
\end{table}
The most obvious use-case for the robot, presenting the exhibits, was also mentioned most often by the visitors, cf.~Table~\ref{tab:pos-scenarios}. However, this was followed by the idea to let Andrea work and interact at the ticket counter of the museum. Interestingly, some visitors explicitly suggested to substitute the audio guides with our robot and even some others wanted to substitute human personnel by robots like Andrea. Only in four cases no positive scenarios came to mind.

% \begin{itemize}
%     \item[\textbf{(5)}] \textbf{What should be urgently improved before such a robot is used?}
% \end{itemize}
\paragraph*{(5) What should be urgently improved before such a robot is used?}

\begin{table}[ht]
    \centering
    \caption{Possible improvements mentioned by the visitors}
    \begin{tabular}{|l|c|}
        \hline
        Suggested improvement & Count\\
        \hline
        multi-language support & 11\\
        improved latency & 9\\
        better non-verbal behavior / facial expressions & 8\\
        better explanation about why the robot is here & 5\\
        access to up-to-date information & 4\\
        slower speech & 3\\
        make it walk & 3\\
        preventing bias / only truthful information & 2\\
        no robot needed, mic and loudspeaker sufficient & 2\\
        nothing concrete mentioned & 2\\
        \hline
    \end{tabular}
    \label{tab:improvements}
\end{table}

Most visitors complained about the need to speak English with Andrea and suggested multi-language support as the most important improvement, cf.~Table~\ref{tab:improvements}. Also, the latency of more than two seconds between question and response, cf.~Section~\ref{sec:conversationAnalysis}, was mentioned by nine visitors. Better non-verbal behavior generation and slower speech were requested more than once and the robot's purpose should be explained better. Those visitors with previous knowledge about ChatGPT suggested access to up-to-date information and were worried that the robot could provide untruthful information.

\subsection{Conversation analysis}
\label{sec:conversationAnalysis}

\subsubsection{Visitor requests}

% \begin{figure*}[ht]
%     \centering
%     \includegraphics[width=1\linewidth]{images/input_topics.png}
%     \includegraphics[width=1\linewidth]{images/output_topics.png}
%     \caption{The topics visitors talked about to the robot (top) and the topics the robot talked about to the visitors (bottom).}
%     \label{fig:topicclusters}
% \end{figure*}

\begin{table}[ht]
\centering
\caption{The topics visitors talked about to the robot}
\label{tab:input_clusters1}
\begin{tabular}{|l|l|rr|}
\hline
Category                               & Cluster name                & \multicolumn{2}{r|}{Requests}                    \\ \hline
\multirow{6}{*}{Personal quest.}     & name\_your\_what\_is        & \multicolumn{1}{r|}{460} & \multirow{6}{*}{1608} \\ \cline{2-3}
                                       & where\_what\_here\_doing    & \multicolumn{1}{r|}{388} &                       \\ \cline{2-3}
                                       & how\_are\_you\_hi           & \multicolumn{1}{r|}{363} &                       \\ \cline{2-3}
                                       & old\_how\_are\_you          & \multicolumn{1}{r|}{143} &                       \\ \cline{2-3}
                                       & color\_favorite\_food\_like & \multicolumn{1}{r|}{102} &                       \\ \cline{2-3}
                                       & who\_long\_did\_when        & \multicolumn{1}{r|}{152} &                       \\ \hline
\multirow{3}{*}{Convers. phrase} & hello\_come\_we\_going      & \multicolumn{1}{r|}{430} & \multirow{3}{*}{818}  \\ \cline{2-3}
                                       & thank\_day\_much\_nice      & \multicolumn{1}{r|}{295} &                       \\ \cline{2-3}
                                       & bye\_goodbye\_okay\_thank   & \multicolumn{1}{r|}{93}  &                       \\ \hline
\multirow{2}{*}{Context}               & the\_car\_mercedes\_is           & \multicolumn{1}{r|}{469} & \multirow{2}{*}{580}  \\ \cline{2-3}
                                       & museum\_this\_the\_in       & \multicolumn{1}{r|}{111} &                       \\ \hline
Language                               & speak\_german\_can\_you     & \multicolumn{2}{r|}{462}                         \\ \hline
Skill                                  & plus\_100\_times\_what      & \multicolumn{2}{r|}{54}                          \\ \hline
\multirow{2}{*}{Other}                 & ciao\_no\_cheers\_im        & \multicolumn{1}{r|}{174} & \multirow{2}{*}{914}  \\ \cline{2-3}
                                       & can\_me\_you\_to            & \multicolumn{1}{r|}{740} &                       \\ \hline
\end{tabular}
\end{table}

In total 4524 input requests were parsed from the log files, as described in Section~\ref{sec:approach-conversation}. The STT model generated non-ascii characters in 88 requests, which
%Although used in \textit{translate} mode, these characters mainly belong to languages other than English, some characters are emojis or other special characters. These requests 
are excluded from further analysis. The remaining 4436 requests are automatically clustered into 15 clusters and these clusters are manually grouped into six categories as shown in Table~\ref{tab:input_clusters1}.

Most requests are categorized as "personal questions", asking the robot for its name, its origin, its purpose (at the museum), how it is doing, its age and its favorites, esp.~color and food. The category "conversational phrase" contains mainly greetings, thanks and sendoffs. Questions regarding the museum, its exhibits and about cars are categorized as "context", since cars are the main theme of the museum. The "language" category contains requests to speak in a language other than English (most often German). The skill category mainly contains requests to solve math equations. Finally, all outliers as well as a quite big cluster with less clearly assignable requests constitute the category named "other".

There are some clusters partially overlapping the categories. % mainly because of the automated clustering of requests. 
As an example, there is the cluster \textit{where\_what\_here\_doing} with many location related questions. It was categorized as "personal questions" since questions like \textit{Where are you from?} are common in this cluster. However, there are also questions in this cluster like \textit{Do you know where you are?}, in which case the visitor tries to asses the robot's skills. 
There are also requests that are not clearly assignable to a category. For example, in the "context" category it is not clear, if a visitor asks questions about the museum out of interest or rather to assess the robot's skills by asking for already known information. 
%The same problem arises for requests to communicate in a different language. A user might want to actually communicate in a language other than English or might just want to test the robot's capabilities. Thus, these categories are kept separate and the skill category almost only contains simple math questions, for which it is obvious that the robot's capabilities are explored. 
The \textit{can\_me\_you\_to} cluster contains similarly problematic examples. For example, \textit{Can you see me?} belongs to the "skill" category, but the same cluster also contains requests like \textit{Can you tell me a joke please?}, when the visitor probably seeks to be entertained, and some incomplete requests like \textit{You can see the}, that cannot be categorized at all.

\subsubsection{Robot responses}

\begin{table}[ht]
\centering
\caption{The topics the robot talked about to the visitors}
\label{tab:input_clusters2}
\begin{tabular}{|l|l|rr|}
\hline
Category                               & Cluster name                & \multicolumn{2}{r|}{Responses}                   \\ \hline
\multirow{4}{*}{Personal inform.}  & robot\_android\_an\_you     & \multicolumn{1}{r|}{1111}& \multirow{4}{*}{1507} \\ \cline{2-3}
                                       & age\_2022\_in\_was          & \multicolumn{1}{r|}{150} &                       \\ \cline{2-3}
                                       & name\_Andrea\_my\_is           & \multicolumn{1}{r|}{127} &                       \\ \cline{2-3}
                                       & japan\_alab\_in\_company    & \multicolumn{1}{r|}{119} &                       \\ \hline
\multirow{4}{*}{Convers. phr.} & hello\_today\_how\_assist   & \multicolumn{1}{r|}{689} & \multirow{4}{*}{1137} \\ \cline{2-3}
                                       & if\_free\_feel\_ask         & \multicolumn{1}{r|}{211} &                       \\ \cline{2-3}
                                       & day\_goodbye\_great\_have   & \multicolumn{1}{r|}{153} &                       \\ \cline{2-3}
                                       & would\_like\_there\_anything& \multicolumn{1}{r|}{84}  &                       \\ \hline
\multirow{3}{*}{Context}               & the\_robot\_android\_as     & \multicolumn{1}{r|}{430} & \multirow{3}{*}{959}  \\ \cline{2-3}
                                       & museum\_the\_mercedes\_benz       & \multicolumn{1}{r|}{302} &                       \\ \cline{2-3}
                                       & the\_mercedesbenz\_first\_of         & \multicolumn{1}{r|}{227} &                       \\ \hline
Language                               & speak\_german\_yes\_can     & \multicolumn{2}{r|}{318}                         \\ \hline
\multirow{2}{*}{Apologies}             & ai\_language\_model\_an     & \multicolumn{1}{r|}{334} & \multirow{2}{*}{533}  \\ \cline{2-3}
                                       & please\_your\_clarify\_context & \multicolumn{1}{r|}{199} &                     \\ \hline
Other                                  & \begin{tabular}[c]{@{}l@{}}scientists\_atoms\_\\ everything\_trust\end{tabular} & \multicolumn{2}{r|}{1}                    \\ \hline

\end{tabular}
\end{table}
In case of an exception neither input nor output were logged. Thus, the same amount of output responses as input requests are parsed from the log files, namely 4524. Although prompted to respond in plain English, ChatGPT's output contains non-ascii characters in 69 responses. They mainly belong to languages other than English and few of them are emojis (smiling, waving and music notes). The 15 automatically created clusters are manually grouped into six categories, as shown in Table~\ref{tab:input_clusters2}.

The categories of the outputs mainly correspond to the input ones, except that responses to math questions do not turn out as their own cluster. Instead, ChatGPT's tendency to apologize results in a new category that consists of two clusters. In case of the bigger one, labeled \textit{ai\_language\_model\_an}, ChatGPT ignored the system prompt, apologized for the confusion, and stated that it was an AI language model (not an android robot). The second cluster mainly contains apologies for not understanding the visitor's request. 
%Again, there is some overlap between clusters, e.g., responses in the biggest cluster often start with \textit{Hello! My name is...}, where the greeting would belong to conversational phrases, but the introduction provides personal information. 
It is unclear, why there is one outlier remaining after the clustering. Its content is \textit{Why don't scientists trust atoms? Because they make up everything!}. The same joke was told by ChatGPT 15 times during this study. However, in other cases it was surrounded by other texts, like \textit{Sure, here's a joke for you:} and most often clustered into the \textit{ai\_language\_model\_an} cluster.

\subsubsection{Other logs}

\begin{table}[ht]
\caption{Descriptive statistics of processing times per ML model (left) with variables these times depend on (right)}
\label{tab:processing_times}
\begin{tabular}{lrrrr|rrr} 
& \multicolumn{4}{l|}{Processing times (s)} & \multicolumn{3}{l}{Dependencies}\\ \hline
     & STT      & Chat    & TTS    & Lip   & In(s) & Out(c) & Out(s) \\ 
mean & 1.70     & 1.38    & 0.95   & 2.65       & 2.56      & 125.59    & 8.04 \\ 
std  & 1.72     & 0.43    & 0.41   & 2.35       & 2.06      & 70.07     & 3.97 \\ 
min  & 0.60     & 0.47    & 0.25   & 0.23       & 0.02      & 3         & 1.15 \\ 
25\% & 0.87     & 1.08    & 0.69   & 1.22       & 1.42      & 73        & 5.20 \\ 
50\% & 0.97     & 1.34    & 0.91   & 2.21       & 2.24      & 118       & 7.53 \\ 
75\% & 2.34     & 1.61    & 1.13   & 3.64       & 3.29      & 168       & 10.26 \\ 
max  & 29.61    & 8.01    & 8.11   & 80.44      & 26.53     & 1349      & 66.54 \\ 
\end{tabular}
\end{table}

Table~\ref{tab:processing_times} shows descriptive statistics of the logged processing times for each of the four steps in the software pipeline. These times depend on variables for which statistics are provided, too,
%Also, the statistics for some dependent variables are provided, 
with \textit{In(s)} being the duration of the recorded input audio in seconds, \textit{Out(c)} the length of the output generated by ChatGPT in characters, and \textit{Out(s)} the duration of the synthesized speech in seconds. 
%However, the time required by LLMs like ChatGPT to generate an answer also depends on the input they need to process, but we did not log the current length of the message stack for each request. 
On average generating mouth movements for a speech signal takes the longest with $2.65s$. Thus, to improve the latency of the whole system, as requested in the interviews (cf. Table~\ref{tab:improvements}), this step of the pipeline will be improved next.
Some requests took exceptionally long to process, as indicated by the gap between maxima and the third quartile. Those are usually the first runs of the pipeline after startup of the Jetson Orin each morning and are part of test runs, not actual user interactions. Therefore, the median values (50\%) should be preferred over the means to measure the system's latency. As such, the total median response time from end of recording until the first spoken word is 5.43 seconds.

%Analyzing \textbf{errors} the system turns out to run quite robust, with only 163 logged exceptions over the whole duration of this study.
With only 113 logged exceptions during all six days, the system has been running quite robustly. All of the exceptions are related to ChatGPT and 50 are request timeouts that we cannot influence in any way. The remaining 63 exceptions were caused by exceeded context lengths%, i.e., no one manually resetting the dialog early enough
. In all cases Andrea responded with: "\textit{I am sorry my artificial intelligence module seems to be a little bit glitchy at the moment. Give me a second to sort my circuits and retry your request.}" However, the system continued to work as intended, after the ChatGPT context was reset manually.

During the four days when this particular information was logged, Andrea got interrupted 640 times while talking. This suggests, that the quality of ChatGPT's responses could be improved. It might make conversations more interesting, if the robot was more context-aware, less polite, and apologized less often, e.g., by adjusting the system prompt.

\subsection{Limitations}
Both methods to acquire information have their limitations. The logs are technically limited, as described, since the automated transcripts may differ from the actual spoken visitors' requests and the automated clustering rather results in an overview of the data but may be imperfect on a single request level. Interviews are known to be limited regarding their quantity and they might be subject to a selection bias towards people interested in the robot \cite{belpaeme2020advice}. We tried to mitigate this effect by also motivating visitors to participate in an interview, who had been reluctant to interact with the robot.

\section{Discussion \& conclusion}
\label{sec:discussion}

We presented the results of an exploratory study of the fully autonomous, android robot Andrea being placed in a public museum in Germany for six consecutive days. The interview data and the cluster analysis of the conversation log files show corresponding results.
For example, many interviewed visitors suggested multi-language support as an improvement and this was also requested directly during conversations with the robot. The \textit{Context} category, identified in the logs, supports the most common positive scenario retrieved from the interviews \textit{presenting knowledge about exhibits} as the most important one. The biggest category extracted from the log files, \textit{personal questions}, reflects multiple of the \textit{Reasons (not) to interact} stated during the interviews: people, who \textit{did not know what to ask}, probably tend to simply say \textit{hello}. % or ask simple questions like \textit{How are you?}. Further questions like \textit{Where do you come from?} may have been asked out of \textit{curiosity}.

Maximum response times of one second are preferred by users of conversational robots and longer response times should be moderated using conversational fillers \cite{shiwa_how_2008}. Our system's median response time, however, is 5.43 seconds. The robot's listening and thinking gestures might mediate some negative effects, but improved response times were also requested by many interviewees.
%The areas of improvement are also similar to those found in the study with the android robot Nadine employed in a work-place \cite{vishwanath_humanoid_2019}. 

The general attitude towards our robot was more positive than that reported in previous research with android robots in public spaces, cf.~\cite{vishwanath_humanoid_2019,becker-asano_exploring_2010}. Admittedly, similarly to the previous work our results might be influenced by the novelty effect \cite{belpaeme2020advice}, because visitors experienced Andrea only once.
% Still most visitors visit a musuem only once, thus android robots might retain their novelty for some time in this specific context.

In conclusion, our exploratory study confirmed that in a non-critical scenario in a museum an android robot is not perceived as uncanny but as an object of curiosity by most visitors, even when it is run in a fully autonomous fashion and not being tele-operated. Our modification of perceived gender was either not successful or had no significant effect.
Additionally, we gained valuable insights on how an android robot can be employed in a museum, how visitors relate the robot to employees and audio-guides already in place (cf. \cite{velentza_museum_2020}) and what technical improvements are required to make it even more useful to visitors. 
Next, we began to improve the current system to meet these requirements: We provided information about exhibits using RAG (Retrieval Augmented Generation) in combination with ChatGPT 4.1, we replaced VITS with the multilingual XTTS\cite{casanova24_interspeech} system and we sped up response times using FaceXHubert\cite{FaceXHuBERT_Haque_ICMI23} as the base for lip-syncing. We also improved Andrea's non-verbal behaviors by simulating and expressing emotions using WASABI \cite{becker-asano.2014} in the aim to increase its overall believability \cite{Becker-Asano+Wachsmuth.2010}.

These improvements will soon be tested during another exhibition of Andrea at the same museum where it then will fulfill the role of a multi-lingual information agent.

\section*{Acknowledgment}
We thank the Mercedes-Benz Heritage GmbH and the German Research Foundation for their financial support. 

\bibliographystyle{IEEEtran}
\bibliography{references}

\end{document}